\documentclass{article}

\usepackage{microtype}
\usepackage{graphicx}
\usepackage{subcaption}
\usepackage{booktabs} 

\usepackage{hyperref}

\usepackage[preprint]{icml2026}

\usepackage{amsmath}
\usepackage{amssymb}
\usepackage{mathtools}
\usepackage{amsthm}
\usepackage{multirow}%
\usepackage{array}

\usepackage[capitalize,noabbrev]{cleveref}

\theoremstyle{plain}

\theoremstyle{definition}

\theoremstyle{remark}

\usepackage[textsize=tiny]{todonotes}

\icmltitlerunning{Submission and Formatting Instructions for ICML 2026}

\begin{document}

\twocolumn[
  \icmltitle{Physics-Guided Multimodal Transformers are the Necessary Foundation for the Next Generation of Meteorological Science}



  \icmlsetsymbol{equal}{*}

  \begin{icmlauthorlist}
    \icmlauthor{Jing Han}{bupt}
    \icmlauthor{Hanting Chen}{huawei}
    \icmlauthor{Kai Han}{huawei}
    \icmlauthor{Xiaomeng Huang}{thu}
    \icmlauthor{Wenjun Xu}{bupt}
    \icmlauthor{Dacheng Tao}{ntu}
    \icmlauthor{Ping Zhang}{bupt2}
  \end{icmlauthorlist}

  \icmlaffiliation{bupt}{School of Artificial Intelligence, Beijing University of Posts and Telecommunications}
  \icmlaffiliation{huawei}{Huawei Noah's Ark Lab}
  \icmlaffiliation{thu}{Department of Earth System Science, Tsinghua University}
  \icmlaffiliation{ntu}{College of Computing \& Data Science, Nanyang Technological University}
  \icmlaffiliation{bupt2}{State Key Laboratory of Networking and Switching Technology, Beijing University of Posts and Telecommunications}

  \icmlcorrespondingauthor{Firstname1 Lastname1}{first1.last1@xxx.edu}
  \icmlcorrespondingauthor{Firstname2 Lastname2}{first2.last2@www.uk}

  \icmlkeywords{Machine Learning, ICML}

  \vskip 0.3in
]



\printAffiliationsAndNotice{}  

\begin{abstract}
	This position paper argues that the next generation of artificial intelligence in meteorological and climate sciences must transition from fragmented hybrid heuristics toward a unified paradigm of physics-guided multimodal transformers. While purely data-driven models have achieved significant gains in predictive accuracy, they often treat atmospheric processes as mere visual patterns, frequently producing results that lack scientific consistency or violate fundamental physical laws. We contend that current ``hybrid'' attempts to bridge this gap remain ad-hoc and struggle to scale across the heterogeneous nature of meteorological data ranging from satellite imagery to sparse sensor measurements. We argue that the transformer architecture, through its inherent capacity for cross-modal alignment, provides the only viable foundation for a systematic integration of domain knowledge via physical constraint embedding and physics-informed loss functions. By advocating for this unified architectural shift, we aim to steer the community away from ``black-box'' curve fitting and toward AI systems that are inherently falsifiable, scientifically grounded, and robust enough to address the existential challenges of extreme weather and climate change.
\end{abstract}

\section{Introduction}\label{sec1}

The field of meteorological science, encompassing both weather and climate sciences, presents some of the most intricate and dynamically evolving challenges within contemporary scientific research \cite{edwards2011history,stute2001global}. Accurately understanding and predicting the behavior of atmospheric systems, ocean currents, and climate variables demands the analysis of massive datasets and the integration of diverse observational models. These challenges stem from the inherent variability of weather and climate phenomena, their long-term temporal scales, and the spatially heterogeneous nature of the data involved \cite{bauer2015quiet,abraham2002performance}. 

In light of these complexities, machine learning, and in particular deep learning \cite{lecun2015deep}, has brought transformative advantages to meteorological science \cite{Ren2021DeepLW}. The ability to leverage large-scale computational resources has led to breakthroughs in processing complex datasets, opening possibilities for more accurate simulations where even incremental improvements can lead to profound real-world impacts, from policy decisions to disaster response.

However, we argue that the current trajectory of AI-driven meteorology is reaching a critical crossroads. Purely data-driven approaches, while high in predictive accuracy, often treat atmospheric variables as mere numerical pixels, ignoring the fundamental physical constraints that govern our planet. To address this, the community has turned to hybrid models that combine neural networks with established physical principles \cite{reichstein2019deep,hess2022physically}. Yet, we observe that many current hybrid approaches remain fragmented and ad-hoc, typically relying on task-specific heuristics or simple feature-concatenation. This lack of a unified architecture hinders the field's ability to systematically scale across heterogeneous data sources, such as satellite imagery, sensor networks, and numerical outputs, into a coherent, scientifically consistent model.

\textbf{This position paper argues that the future of meteorological AI must move beyond fragmented hybrid heuristics toward a unified paradigm: Physics-Guided Multimodal Transformers.} We contend that the transformer architecture is not merely another neural backbone, but a unique mathematical bridge capable of aligning heterogeneous observational data with explicit domain knowledge. We advocate for a shift toward frameworks that systematically embed physical laws into the multi-head attention mechanisms and architectural design, rather than treating physical signals as secondary inputs. 

By adopting this multimodal transformer paradigm, the field can transcend ``black-box'' curve fitting and develop foundational tools for next-generation meteorological systems. Such systems will be capable of addressing diverse challenges ranging from extreme weather forecasting to climate change projection with improved reliability, falsifiability, and scientific consistency. In the following sections, we outline three distinct integration mechanisms and identify the necessary steps for the community to realize this unified vision.

\begin{figure*}[htp]
	\centering
	\includegraphics[width=1.0\textwidth]{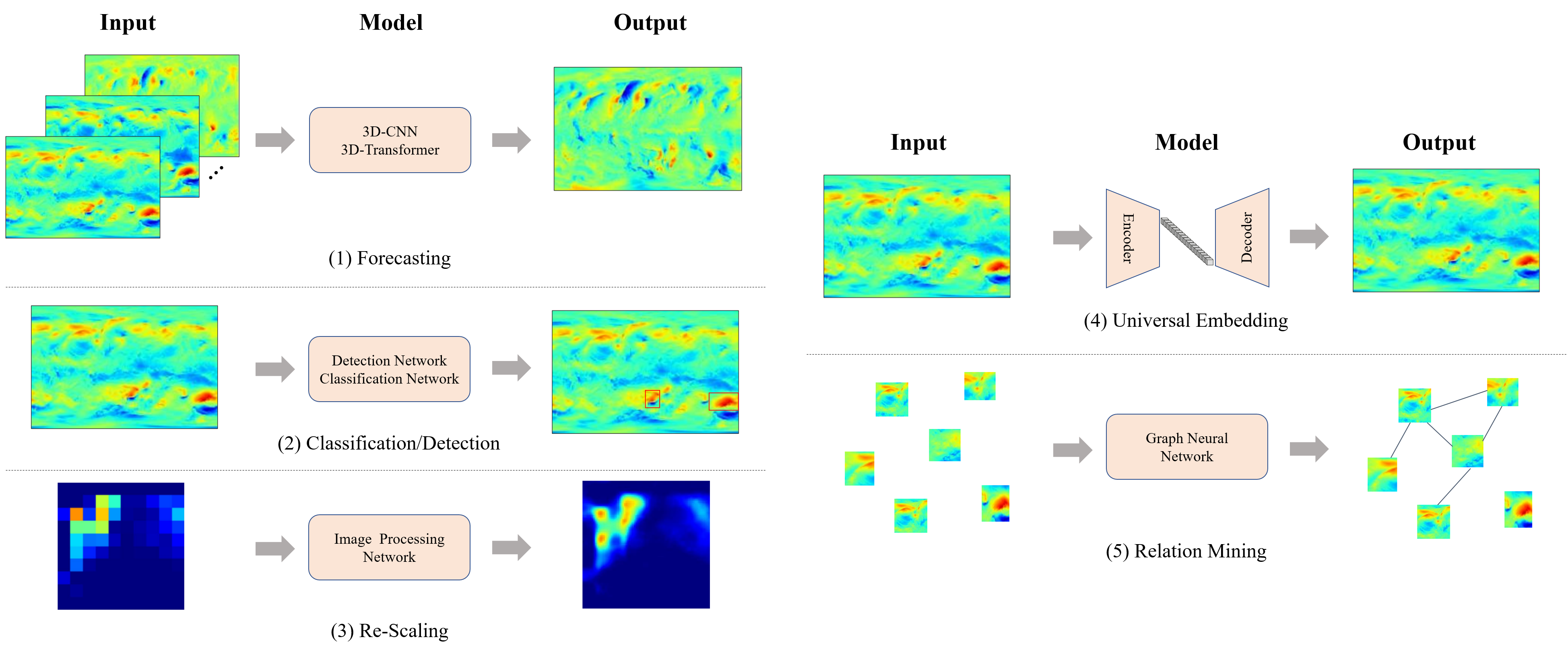}
	\caption{Illustration of the categories of AI models in climate science. The inputs, outputs, and neural network architectures of these models usually vary significantly.}\label{fig:AI}
	\vspace{-1em}
\end{figure*}

\section{State-of-the-art Hybrid Models in Weather and Climate Sciences}\label{sec3}
While AI models, especially those based on deep learning, have shown promise in weather forecast and climate prediction, their reliance on data alone presents significant challenges. These models can produce accurate results, but often suffer from issues related to poor interpretability, limited generalization, and an inability to fully grasp the underlying physical laws governing the atmosphere. In contrast, hybrid models~\cite{reichstein2019deep}, which combine both deep learning and physical modelling, offer a powerful solution to these challenges. By integrating the strengths of data-driven approaches with the robustness of theory-based models, hybrid approaches can leverage the flexibility and adaptability of machine learning while maintaining the interpretability and extrapolation capabilities of physical models.

In this section, we analyze and summarize the current state-of-the-art hybrid models in Table~\ref{tab2}, identifying three key methods for integrating physics priors and constraints into AI models to enhance both their performance and interpretability.

\begin{table*}[h]
	\caption{Hybrid approaches to adding physical priors and representative works.}\label{tab2}
	\centering
	\small
	\begin{tabular}{@{}p{0.16\linewidth}p{0.22\linewidth}p{0.22\linewidth}p{0.32\linewidth}@{}}
		\toprule			
		Approach  & Representative work & Scientific task & Key-points \\
		\midrule
		\multirow{8}{=}{\raggedright Regularization through Loss or Output} 
		& CM2Mc-LPJmL ESM~\cite{hess2022physically} 
		& Earth system models (ESM) simulation 
		& Enforcing a physical constraint on outputs to preserve global precipitation sums \\
		\cmidrule{2-4}
		& NeuralGCM~\cite{kochkov2024neural} 
		& Physical simulation of Earth's atmosphere 
		& Using a combination of three loss functions to encourage accuracy and sharpness while penalizing bias. \\
		\cmidrule{2-4}
		& DeepHybrid~\cite{grover2015deep} 
		& Weather Forecasting 
		& Applying physical laws constraints on the output to be spatially smooth. \\
		\midrule
		\multirow{5}{=}{\raggedright Utilizing Existing Physical Models (PDEs)} 
		& NowcastNet~\cite{zhang2023skilful} 
		& Extreme precipitation nowcasting 
		& Combining physical-evolution schemes and conditional-learning methods \\
		\cmidrule{2-4}
		& ClimODE~\cite{climode} 
		& Climate and weather prediction 
		& Learning a continuous-time PDE model, grounded on continuity equation \\
		\midrule
		\multirow{5}{=}{\raggedright Physics as an Input Signal} 
		& CLLMate~\cite{li2024cllmate} 
		& Weather and Climate Events Forecasting 
		& Knowledge graph and meteorological raster data as inputs of multimodal LLM \\
		\cmidrule{2-4}
		& DUQ~\cite{wang2019deep}, ECMWF Enhancement~\cite{frnda2019weather} 
		& Weather Forecasting 
		& Incorporating prior knowledge from NWP with information fusion mechanism. \\
		\bottomrule
	\end{tabular}
	\vspace{-1em}
\end{table*}

\subsection{Regularization through Loss or Output}
One of the most straightforward ways to incorporate physical knowledge into AI models is by adding physical constraints as regularization terms in the model's loss function. The idea is to guide the model to learn not only from the data but also in a manner consistent with known physical principles. We select two classic cases for a detailed explanation as follows.

\paragraph{Regularization on loss function:} 
General Circulation Models (GCMs) serves as the foundation for simulating and predicting atmospheric dynamics. NeuralGCM~\cite{kochkov2024neural} is a hybrid model for weather and climate by combining GCMs and machine learning. Its inputs include the prognostic variables $X_1, \cdots, X_t$, and other necessary variables $V$ like forcings and noise. NueralGCM will output the prognostic variable tendencies $X_{t+1}$ in the future. The process falls in the scope of Eq.~\ref{eq:forcast}: $X_{t+1} \sim P(X_{t+1} | X_1, \cdots, X_t, V)$. The training object for NueralGCM is a combination of three loss functions. In addition to the standard MSE loss on the main predicted variables, a regularization to force the spectrum to match the training data, and a MSE loss term on the spherical harmonic coefficient's mean magnitude. In general, if the AI model predicts unrealistic velocity profiles or deviates from established physical behaviors like conservation of mass or momentum, a regularization term can be used to adjust the predictions back to plausible values, improving both the accuracy and physical fidelity of the model.

\paragraph{Physical constraint on outputs:} In climate modeling, particularly for precipitation, models aim to capture the complex dynamics of moisture transfer, cloud formation, and precipitation processes. If a model’s predictions for precipitation are unrealistic or violate known physical principles (e.g., precipitation gradients that are overly smooth or unrealistic in regions where intense variability is expected), a physical constraint can be incorporated as an additional layer in the generator network architecture after training to ensure the preservation of the global precipitation sum~\cite{hess2022physically}. The post-processing is performed by a generator, which is a simplification form of Eq.~\ref{eq:rescale}: $X'_{t} \sim P(X'_{t} | X_t)$. A physical constraint to preserve the total global precipitation amount of CM2Mc-LPJmL model is applied on the output:
\begin{equation}
	X''_{t,i} = X'_{t,i}\frac{\sum_{i}^{N}S_{t,i}}{\sum_{i}^{N}X'_{t,i}},
\end{equation}
where $i$ is the grid index, $N$ is the total number of grid points, $S_{t,i}$ is the input of the CM2Mc-LPJmL model and $X''_{t,i}$ is the regularized output. This helps ensure that the model’s output aligns better with the observed spatial variability of precipitation, thereby improving its physical consistency.

\subsection{Utilizing Existing Physical Models}
The physical principles that are widely used in climate models include the law of conservation of mass, the ideal gas law, the first law of thermodynamics and continuity equations~\cite{Ren2021DeepLW}. A powerful method for integrating physics into AI models is to incorporate existing physical models such as partial differential equations (PDEs) that describe fundamental physical processes into the model's architecture. This approach can help the AI model to better respect the underlying dynamics of the physical system it is modeling.

A representative work is climate modeling with PDEs. Climate models often rely on PDEs to describe atmospheric phenomena such as heat transfer, fluid flow, or radiation. A typical example is the continuity equation in fluid dynamics, which describe the movement of quantities over time:
\begin{equation}\label{eq:pde}
	\frac{du}{dt} + \left(v \cdot \nabla u + u \nabla \cdot v\right) = s,
\end{equation}
where $u$ is a quantity such as temperature driven by the flow’s velocity $v$ and sources $s$.
AI models can be integrated with these equations by using Physics-Informed Neural Networks (PINNs), where the neural network learns to approximate solutions to the PDEs. For example, ClimODE~\cite{climode} introduces a neural network to predict flow velocity and utilize the predicted velocity in the continuity equation for weather prediction. 
It is a typical application of Eq.~\ref{eq:forcast}: $X_{t+1} \sim P(X_{t+1} | X_1, \cdots, X_t)$, where $X$ is the quantity variables to be predicted, and $P()$ consists of flow velocity network and the continuity equation (PDE in Eq.~\ref{eq:pde}) is solved to obtain the prediction results. By incorporating PDE equation as part of the model's architecture or training process, the AI model can ensure its predictions are physically consistent with how quantities should propagate through the environment. This approach allows the AI to use the known mathematical description of the process to guide its learning, resulting in more accurate and realistic predictions of quantities changes over time.

\subsection{Physical Knowledge as an Input Signal}
A third method involves feeding physics-based input signals into the AI model. Rather than embedding physical laws directly into the model architecture, this approach provides the model with explicit physical knowledge or prior that are essential for accurate predictions. The prior could include physical laws such as conservation of mass and energy, empirical physical models like fluid dynamics equations, or boundary conditions such as initial atmospheric conditions, which can be used by the model to guide its learning process.

In weather forecast, key physical priors like weather and climate knowledge, as well as common-sense knowledge are used as input features for AI models. For instance, in CLLMate~\cite{li2024cllmate}, a multimodal large language model (LLM) for weather and climate event prediction, the authors propose a framework that uses a LLM to extract information from  environment-focused news articles and build a knowledge graph. This graph, combined with meteorological raster data, is used in the multimodal LLM to predict weather and climate events. By incorporating such input signals into the model, it can learn not just from the raw observational data (e.g., satellite images, sensor data), but also from the underlying physical processes that govern weather dynamics. It can be viewed as an example of Eq.~\ref{eq:class}: $	Y_{t} \sim P(Y_{t} | X_1, \cdots, X_t, S)$,	where $X$ is the raw observational data, $S$ is the physical knowledge, and $Y_{t}$ is weather and climate event prediction results. The multimodal LLM with parameters $\Theta$ performs the whole prediction process.

\section{Limitations and Challenges of Current Hybrid Models}

While hybrid models combining AI and physical modeling have shown promise in addressing challenges in weather forecast and climate prediction, several limitations and challenges remain. These issues hinder the full potential of hybrid models and present significant barriers to their application in real-world scenarios. In this section, we outline the key challenges that current hybrid models face.

\subsection{Lack of Unified Model Architecture}

One of the most significant challenges with current hybrid models is the lack of a unified or standardized model architecture. At present, different AI techniques, such as convolutional neural networks, recurrent neural networks (RNNs), and transformers, are employed in varying configurations depending on the task or domain. This diversity in architecture has led to inconsistent results and a lack of comparability across studies. For instance, CNNs excel at spatial feature extraction, while RNNs are well-suited for sequential data processing, and transformers are particularly effective in capturing long-range dependencies. However, each of these models operates as an isolated component in most hybrid systems, without a standardized method for their integration with physical models.

The absence of a unified framework complicates the process of designing and optimizing hybrid models, making it difficult to establish best practices or identify generalizable principles. A more cohesive, flexible, and standardized approach to hybrid model architecture would facilitate greater integration of AI and physics, enabling the development of models that can be more easily adapted to different applications. Furthermore, without a common framework, the comparison of results across different studies becomes challenging, limiting the potential for cross-domain knowledge transfer and slowing the overall advancement of hybrid modeling techniques.

\subsection{Limited Multi-modal Data Fusion}

Another limitation of many current hybrid models is their reliance on single-modal data inputs. Most hybrid models focus on processing one type of signal or data source, such as atmospheric pressure, temperature, or wind speed, often neglecting the rich variety of data available in real-world settings. This limited focus reduces the model’s ability to fully capture the complexity of natural systems, which typically involve multiple interacting variables from diverse sources.

For example, weather forecast models that only incorporate temperature data may fail to account for the impacts of humidity, air quality, or satellite-derived images, which often contain valuable information about atmospheric conditions. Similarly, climate models that only process physical measurements may miss important signals from remote sensing technologies or socio-economic factors that are critical for long-term projections. The lack of effective multi-modal data fusion constrains the model’s ability to make accurate predictions, particularly in scenarios that involve complex interactions between multiple data sources. A more robust approach to integrating data from different modalities would allow hybrid models to take full advantage of the available information, improving both accuracy and reliability.

\subsection{Challenges in Scaling to Large, Complex Systems}

Scaling hybrid models to large, complex systems is another significant challenge. While hybrid models may perform well for small-scale problems or specific datasets, their effectiveness often diminishes when applied to more extensive and intricate systems, such as global climate models or large-scale weather prediction tasks. These complex systems involve numerous interacting variables, nonlinear relationships, and spatiotemporal dependencies, which are difficult to model accurately using current hybrid approaches. Additionally, the calibration and validation of these models become increasingly challenging as the number of variables and data sources increases, leading to issues with model convergence and reliability.

To overcome these challenges, hybrid models need to be optimized for adapting to complex systems. This requires the development of more sophisticated techniques for handling high-dimensional data, managing nonlinearity, and capturing intricate spatiotemporal dependencies. One promising approach is the integration of deep learning methods, particularly those that can efficiently process large amounts of unstructured data, with traditional physical-based models. This combination can help leverage the strengths of both approaches—deep learning’s ability to model complex, nonlinear relationships and the physical model’s ability to incorporate domain knowledge and respect known conservation laws.

\section{Perspective: Towards a Unified Multimodal Transformer Path}

In this work, we propose to advance climate science through the integration of a physics-guided multimodal transformer framework. This approach aims to address the challenges associated with existing hybrid models by seamlessly combining diverse weather and climate data modalities while embedding critical physical principles. By doing so, we highlight how transformer models can not only handle complex, multi-dimensional atmospheric data but also leverage physical knowledge to enhance model interpretability, accuracy, and scientific validity.

\begin{figure*}[h]
	\centering
	\includegraphics[width=0.8\textwidth]{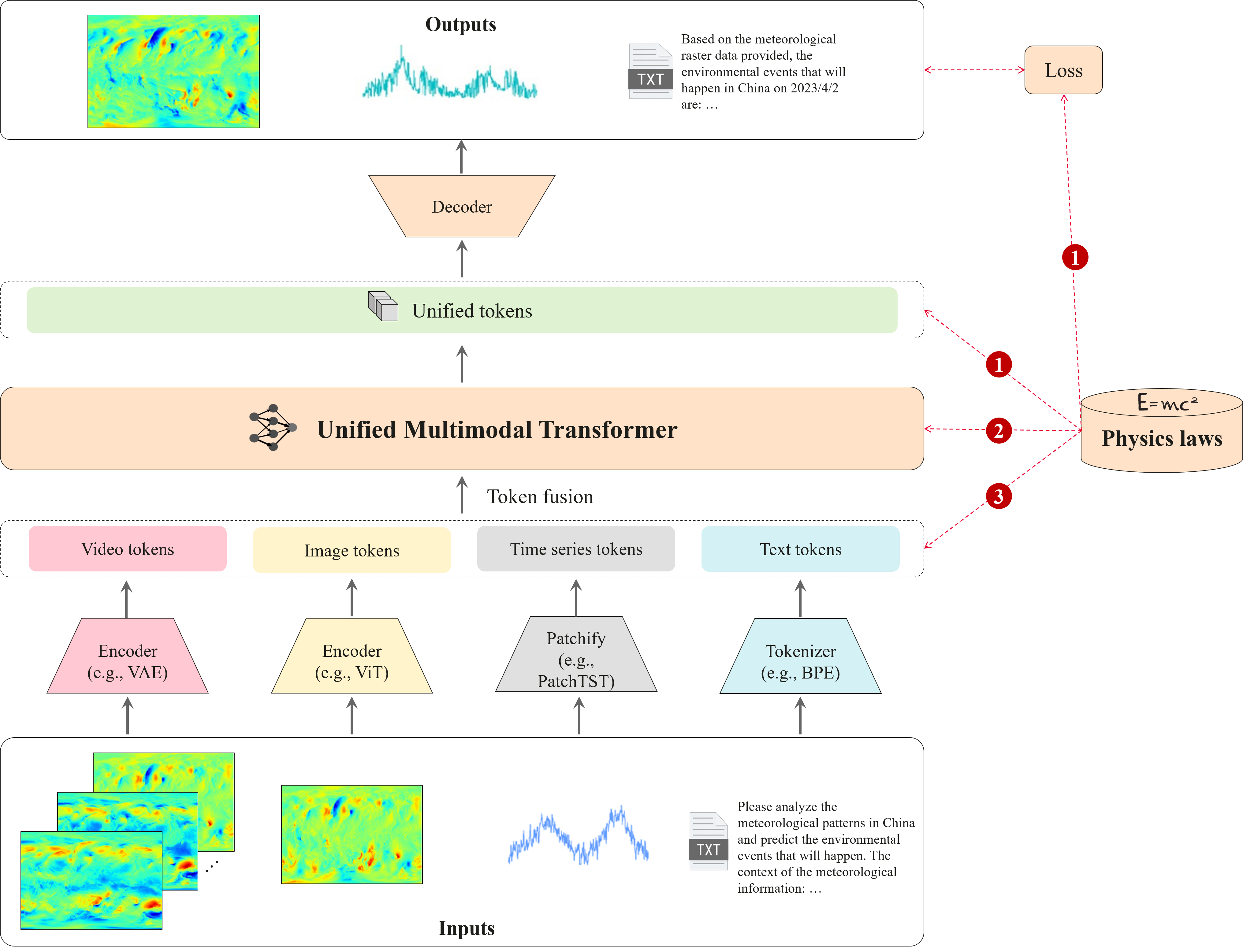}
	\caption{Illustration of the physics-guided multimodal transformer path to weather and climate sciences. The circle labelled 1, 2, 3 represents adding physical knowledge into the output, model and input, respectively.}\label{fig:framework}
	\vspace{-1em}
\end{figure*}

\subsection{Multi-modal Fusion}
In weather and climate sciences, the input signals are diverse, including satellite observation data, ground-based observation data, simulation data, historical time-series data, textual records, and more. These constitute typical multimodal data and can be categorized to image-like data $X_I\in\mathbb{R}^{H\times W}$, spatial-temporal video data $X_V\in\mathbb{R}^{T\times H\times W}$, time series data $X_{S}\in\mathbb{R}^{T}$ and text data $X_{T}\in\mathbb{N}^{T}$, where $T$ is the time length, and $H\times W$ is the spatial size. In order to process these data in a unified transformer, we propose to encode them into tokens with the domain-specific tokenizer or encoder. 

\begin{itemize}
	\item For image-like data \( X_I \in \mathbb{R}^{H \times W} \), a transformer encoder (e.g., ViT~\cite{Dosovitskiy2020AnII}) is employed to extract semantic representations, which are then flattened into a sequence of tokens \( X'_I \in \mathbb{R}^{N \times D} \).
	\item For video-like data \( X_V \in \mathbb{R}^{T \times H \times W} \), the video can be converted into spacetime patches by compressing both the spatial and temporal dimensions into a latent space~\cite{videoworldsimulators2024}. These patches are subsequently flattened into a sequence of tokens \( X'_V \in \mathbb{R}^{N \times D} \) for further processing.
	\item When dealing with time series data \( X_S \in \mathbb{R}^{T} \), the data can be divided into overlapping or non-overlapping patches~\cite{nietime}, which are then linearly projected into the target domain \( X'_S \in \mathbb{R}^{N \times D} \).
	\item For text data \( X_T \in \mathbb{N}^{T} \), a tokenizer such as BBPE~\cite{wang2020neural} is used to convert the text into discrete tokens \( X'_T \in \mathbb{R}^{N \times D} \).
\end{itemize}

The key innovation of our approach lies in the representation and processing of different atmospheric layers as separate modalities. For instance, high-altitude and low-altitude atmospheric data, which are often treated as separate entities in traditional models, are now tokenized and processed independently. This tokenization allows the model to better differentiate between the dynamics at various altitudes and temporal scales, ensuring that the interactions between these layers are captured more effectively. The multimodal transformer framework naturally integrates these different data streams, enabling the model to consider the entire atmospheric system more holistically.

Effective alignment of multimodal data is crucial for achieving accurate predictions in weather and climate sciences. Given the diversity of data types involved, effective alignment ensures that the model can integrate these modalities in a coherent and meaningful way. For simplicity and generalization, we align the different modalities by embedding each modality into a shared latent space, allowing the transformer model to process them in a unified manner. This cross-modal alignment ensures that the model can learn meaningful relationships between modalities, even when they have different temporal and spatial characteristics. Let \( X'_I, X'_V, X'_S, X'_T \) represent the tokenized sequences for the image-like, video-like, time-series, and text data, respectively. The shared embedding space is defined as:
\[
X'_{\text{aligned}} = \mathcal{E}(X'_I, X'_V, X'_S, X'_T),
\]
where \( \mathcal{E} \) is the embedding function that projects each modality into a shared latent space. This cross-modal embedding allows the transformer to process the diverse data streams as a unified input for further processing.

In some cases, we have a primary modality, with other modalities serving as auxiliary inputs to assist the primary modality's task. We can integrate the information from these auxiliary modalities into the primary modality. We use a cross-attention mechanism to combine the auxiliary inputs of different modalities by taking them as keys and values, and the primary modality inputs as queries. Specifically, the output of the cross-attention layer is computed as:
\[
X'_{\text{aligned}} = \text{Softmax}\left(\frac{Q K^T}{\sqrt{d}}\right)V,
\]
where \( Q, K, V \) represent the queries, keys, and values,
\( d \) is the dimensionality of key vectors. The weighted sum of values is then used to produce the output representation, which integrates information from multiple modalities.

\subsection{Unifying Tasks via Next-token Prediction}
At the core of this framework is the transformer model's attention mechanism, which serves as a powerful tool for fusing diverse data sources. Unlike conventional models that may struggle to effectively combine heterogeneous data, the transformer’s self-attention mechanism allows the model to weigh the importance of different data modalities in a context-dependent manner. This dynamic fusion capability enables the model to adapt to varying input conditions, making it highly flexible and capable of handling the complexity inherent in atmospheric science problems. The aligned multi-modal inputs can be represented as a sequence of tokens:
\begin{equation}
	X'_{\text{aligned}} = [\mathbf{x}_1, \mathbf{x}_2, \cdots, \mathbf{x}_N],
\end{equation}
where $\mathbf{x}_i\in\mathbb{R}^{d}$, $i=1,2,\cdots,N$ and $N$ is the number of tokens.

The model is trained to predict the next token in the sequence, effectively learning to generate the most contextually appropriate token based on the preceding tokens.
\begin{equation}
	\mathbf{x}_{\tau+1} \sim P_{\Theta}(\mathbf{x}_1 | \mathbf{x}_1, \mathbf{x}_2, \cdots, \mathbf{x}_{\tau}),
\end{equation}
where $\Theta$ is weights of the transformer model, and $\mathbf{x}_{\tau}$ is the predicted token from the previous tokens and these predicted tokens will form the results we need, such as forecasts for future weather or high-resolution climate data. This next-token prediction task allows the transformer to capture long-range dependencies and complex relationships between the multi-modal inputs, making it well-suited for atmospheric science applications where interactions between various data sources are often non-linear and span across multiple time scales.

\subsection{Physical Priors Injection}
Moreover, we propose to guide the transformer with physical principles such as conservation laws, thermodynamic constraints, and atmospheric dynamics. By embedding these physical priors into the transformer’s learning process, we ensure that the model remains consistent with known physical behaviors, thus improving its generalization capabilities. This guidance is particularly crucial when the model is confronted with novel, unseen conditions, as it enables the model to extrapolate beyond the training data while staying grounded in the laws of physics. As shown in Fig.~\ref{fig:framework}, physical knowledge (denoted as \(\mathcal{K}\)) can be merged into the outputs, model or inputs respectively:
\begin{equation}
	\mathbf{x}_{\tau+1} = f(\mathcal{K}, P_{\Theta,\mathcal{K}}(\mathbf{x}_1 | \mathcal{K}, \mathbf{x}_1, \mathbf{x}_2, \cdots, \mathbf{x}_{\tau})),
\end{equation}
where $f$ is the function to refine outputs with the physical knowledge \(\mathcal{K}\).

Next, we will explain the specific process of injecting physical priors into the outputs, model, and inputs, provide examples for analysis.
\begin{itemize}
	\item Injecting into the outputs (the circle labelled 1 in Fig.~\ref{fig:framework}): Integrating physical priors $\mathcal{K}_{out}$ into the model's initial outputs $\mathbf{x}'_{\tau+1}$ involves adding regularization terms to the loss function or imposing specific constraints on the outputs to ensure physical consistency. For instance, a common approach is to introduce a regularization term that penalizes deviations from known physical laws, such as conservation of mass or energy. Alternatively, we can directly constrain the output to satisfy certain physical bounds: $\mathbf{x}_{\tau+1} = f(\mathcal{K}, \mathbf{x}'_{\tau+1})$, such as limiting temperature predictions within realistic ranges based on historical data or physical models. These constraints guide the model towards solutions that are not only accurate in terms of data fitting but also adhere to fundamental scientific principles, improving the interpretability and reliability of the predictions.
	\item Injecting into the model (the circle labelled 2 in Fig.~\ref{fig:framework}): Integrating Physical Priors into the model involves incorporating physical equations, such as PDEs, directly into the model architecture. One effective approach is the use of Physics-Informed Neural Networks (PINNs), where the neural network is trained not only to minimize the data-driven loss but also to satisfy the governing PDEs that describe the underlying physical processes. This allows the model to leverage both data and physical knowledge, leading to more accurate and physically consistent predictions, especially in scenarios where data is sparse or noisy.
	\item Injecting into the inputs (the circle labelled 3 in Fig.~\ref{fig:framework}): Integrating physical priors $\mathcal{K}_{in}$ into the model's raw inputs $X_{in}$ involves incorporating meteorological knowledge or historical data as additional input signals to guide the model. For example, we can introduce weather patterns, such as wind speed, humidity, or pressure distributions, derived from historical climate data or scientific literature, as supplementary features concatenated with the raw inputs $[X_{in}, \mathcal{K}_{in}]$. These physical inputs provide the model with prior knowledge of atmospheric conditions, enabling it to make more informed predictions. By including these established physical signals, the model is better equipped to understand the underlying dynamics of climate systems, leading to improved accuracy and consistency in its outputs, particularly in forecasting and long-term climate projections.
\end{itemize}

This unified multimodal transformer approach is expected to provide significant benefits across various meteorological science tasks. For instance, in weather forecast, the multimodal transformer can seamlessly integrate real-time observational data with numerical simulations, resulting in more accurate and timely forecasts. In numerical super-resolution, the model effectively reconstructs high-resolution climate data from coarser inputs, enhancing model performance for long-term climate projections. Finally, in climate recognition, the transformer model excels at identifying subtle yet critical patterns within climate datasets, helping to detect emergent behaviors crucial for understanding climate change. 

\subsection{Scalability in Large-Scale Systems}
Large-scale and complex systems, such as global climate models~\cite{stute2001global}, form the cornerstone of modern atmospheric research. These models integrate vast amounts of observational and simulated data to study Earth’s climate, making them critical tools for understanding phenomena ranging from regional weather patterns to global climate change. However, their computational demands, complex parameterizations, and inherent uncertainty present ongoing challenges.

By employing the proposed multimodal transformer framework, we can substantially improve numerical models in several ways. First, these methods offer a pathway to better emulate the underlying physics or dynamics within the numerical models, enabling more accurate representation of key processes. Second, they allow for more effective parameterization. Complex sub-grid processes such as cloud microphysics or ocean-ice interactions can be modeled more reliably by integrating data-driven approaches that respect known physical constraints. This not only enhances the fidelity of numerical simulations but also reduces the need for manually calibrated empirical parameters, leading to more consistent and interpretable outcomes.

Another crucial aspect of scalability lies in the adaptability of the proposed path. As a pretrained foundation model, it can be fine-tuned for various tasks without requiring extensive retraining. For example, after training on a global dataset, the model can be adapted to regional scales or specific phenomena, such as monsoon dynamics or Arctic sea ice conditions. This ability to efficiently fine-tune on new datasets or emerging scientific challenges makes the model highly versatile, allowing researchers to address a wide range of atmospheric science problems with reduced computational effort and time.

Finally, the scalability of the proposed approach extends to the realm of large-scale computation and inference. By leveraging parallel computing architectures, the model can be trained and deployed at scale, significantly accelerating both the training and inference phases. Advanced parallelization strategies, such as model and data parallelism, enable efficient utilization of modern high-performance computing infrastructures. This not only shortens model development cycles but also allows for real-time or near-real-time predictions in operational settings. Furthermore, optimized inference pipelines ensure that large ensembles of predictions can be generated quickly, enabling researchers to explore a wide range of scenarios and to perform more robust uncertainty quantification.

\section{Alternative Views}\label{sec:alternative_views}

While we advocate for the integration of physical laws into multimodal transformers, we acknowledge and address two prominent alternative perspectives in the current machine learning community.

\subsection{The Symmetry of Scaling: Can Data Alone Supplant Explicit Physics?}
A significant counter-argument, rooted in the Scaling Law hypothesis \cite{kaplan2020scaling}, suggests that sufficiently large models trained on massive datasets can implicitly ``discover'' physical principles without explicit guidance~\cite{lam2022graphcast}. Proponents of this view might argue that hard-coding physical constraints acts as a form of inductive bias that may ultimately limit the model's capacity to represent complex, non-linear atmospheric dynamics that are not yet fully captured by classical equations.

\textbf{Our Response:} While scaling has led to impressive performance, ``implicit discovery'' of physics is often fragile. In scenarios involving out-of-distribution events, such as unprecedented extreme weather driven by climate change, purely data-driven models frequently produce physically impossible outputs (e.g., negative precipitation or violations of mass conservation). Explicit physics-guidance ensures that the model remains within the manifold of reality, providing a safety floor that scaling alone cannot guarantee when historical data is no longer a reliable predictor for future.

\subsection{The Computational Trade-off: Accuracy vs. Physical Consistency}
Another credible alternative view posits that the integration of physical constraints, especially through complex loss functions or modified attention kernels, introduces significant computational overhead and optimization challenges. Critics may argue that in operational forecasting, where inference speed is paramount, a faster, ``approximately correct'' data-driven model is more practically valuable than a slower, ``physically perfect'' transformer.

\textbf{Our Response:} We contend that the perceived trade-off between speed and consistency is a false dichotomy. By leveraging the multimodal transformer's ability to fuse data efficiently, physical constraints can be used to \textit{reduce} the search space during training, potentially leading to faster convergence. Furthermore, our proposed paradigm advocates for ``physics-in-architecture'' (e.g., tailored encoding) rather than merely adding expensive differential equation solvers to the inner loop of inference. The goal is to achieve ``Physics-by-Design'', which maintains operational efficiency while ensuring scientific reliability.

\section{Conclusion and Call to Action}\label{sec:conclusion}

This paper has argued that the current fragmented approach to meteorological AI is hitting a wall of scientific unreliability. We contend that the path forward lies in the adoption of a physics-guided multimodal transformer framework as the unified foundation for the field. By moving beyond ``black-box'' curve fitting and toward architectures that treat physical laws as structural imperatives, the machine learning community can provide the reliability and scientific consistency that critical climate policy and disaster response demand.

To realize this vision, we propose the following call to action for the meteorological AI community:

\textbf{1) Develop Physics-Consistent Benchmarks:} We call for a transition from standard error metrics (e.g., MSE) toward evaluation suites that explicitly penalize violations of physical conservation laws. High-quality, multimodal datasets must be co-designed by ML researchers and meteorologists to ensure that accuracy is always grounded in physical reality.
	
\textbf{2) Prioritize ``Physics-by-Design'':} We advocate for a shift from treating physics as secondary input signals to embedding domain knowledge directly into transformer structures. Future research should prioritize architectures where physical constraints—such as fluid dynamics or thermodynamic limits—are integrated into attention mechanisms and encoding schemes, making consistency an inherent property of the model.
	
\textbf{3) Ensure Operational Robustness and Efficiency:} For these advanced frameworks to be viable, they must maintain physical plausibility even under data-sparse conditions (e.g., sensor failure). Furthermore, we urge the community to integrate model optimization techniques (such as quantization and distillation) to ensure that scientific rigor does not come at the cost of the real-time inference speed required for operational forecasting.

In conclusion, the success of AI in weather and climate sciences depends on its alignment with the immutable laws of physics. By guiding the next generation of multimodal transformers with these principles, we can ensure that AI serves as a scientifically robust pillar for understanding and protecting our planet's future.

\bibliography{sn-bibliography}
\bibliographystyle{icml2026}

\newpage
\appendix
\onecolumn
\section{Categories of AI Models in Weather and Climate Sciences}

AI models have begun to make significant strides in the field of weather and climate science. State-of-the-art algorithms in this domain can be broadly categorized into five main types, each addressing different aspects of weather and climate predictions along with their analysis.

\subsection{Forecasting Models}
Forecasting models in weather and climate sciences predict future meteorological conditions using historical data. Convolutional Neural Networks (CNNs) and Graph Neural Networks (GNNs) have been widely used for their ability to capture spatial dependencies and model complex meteorological relationships. CNNs are effective in identifying spatial patterns in variables like temperature and pressure, while GNNs handle dynamic, grid-based weather system data across regions~\cite{sonderby2020metnet, lam2022graphcast}. For instance, MetNet combines CNNs, LSTMs, and autoencoders to capture multi-scale spatial and temporal features for short-term forecasts~\cite{sonderby2020metnet}. Similarly, GraphCast employs GNNs within an ``encode-process-decode'' framework to model intricate spatiotemporal relationships, enhancing flexibility and accuracy over traditional approaches~\cite{lam2022graphcast}. 

However, CNN and GNN-based models often fail to capture long-range temporal dependencies and global interactions crucial for accurate climate forecasting. Transformer-based models, leveraging attention mechanisms, address these limitations by focusing on relevant features across space and time~\cite{bi2022pangu,pathak2022fourcastnet,sun2024fuxi}. Notable examples include PanGu Weather~\cite{bi2022pangu}, which uses a 3D-Swin Transformer for spatiotemporal data processing, and FourCastNet~\cite{pathak2022fourcastnet}, integrating adaptive Fourier neural operators with vision Transformers for high-resolution data. These models better capture complex dependencies and model both local and global features effectively. These developments indicate a shift towards more scalable and flexible solutions, improving predictions of extreme weather events and long-term climate changes.

To optimize these models, we examine the classical objective function used in forecasting tasks. For example, the objective for PanGu Weather~\cite{bi2022pangu} is:
\begin{equation}
	\mathcal{L} = \text{MSE}(X_{t+1}, f(X_{t}, \theta)),
\end{equation}
where \(X_{t+1}\) is the atmospheric state at \(t+1\), and \(f(X_{t}, \theta)\) is the model's prediction based on parameters \(\theta\). This minimizes the mean squared error (MSE) between predicted and actual states, a common approach in supervised weather forecasting.

More generally, forecasting can be viewed as a next-token prediction problem:
\begin{equation}\label{eq:forcast}
	X_{t+1} \sim P(X_{t+1} | X_1, \cdots, X_t),
\end{equation}
where \(X_t\) represents 3D atmospheric data at time \(t\). This aligns with sequence modeling tasks, enabling transformer-based architectures to refine temporal dependencies continuously. In summary, framing climate forecasting as sequential prediction enhances weather and climate models, extending their applicability to various domains requiring robust sequential prediction capabilities.

\subsection{Classification and Detection Models}
Classification and detection models in weather and climate sciences identify specific phenomena, such as extreme weather events, by classifying or localizing occurrences like storms and heatwaves. These models improve localization accuracy and address limited labeled data by extracting meaningful patterns from historical data for future predictions~\cite{liu2016application, racah2017extremeweather, chen2019hybrid}. Deep learning architectures, particularly CNNs, have been widely adopted for climate pattern detection and extreme event classification~\cite{liu2016application, racah2017extremeweather}. Several works combining CNNs with recurrent structures like LSTMs capture the spatio-temporal dynamics of events such as typhoons~\cite{chen2019hybrid}.

For classification tasks, the objective function typically uses cross-entropy loss:
\begin{equation}
	\mathcal{L} = - \sum_{i=1}^{N} \overline{y}_t \log(f(X_t,\theta)),
\end{equation}
where \( \overline{y}_t \) is the true label and \( f(X_t,\theta) \) is the predicted probability at time \( t \). This minimizes the cross-entropy loss to maximize the likelihood of correct class predictions. Detection tasks involve both classification and localization, combining cross-entropy loss with localization loss such as mean squared error.

These tasks can be generalized as next-token prediction problems:
\begin{equation}\label{eq:class}
	Y_t \sim P(Y_t \mid X_1, \cdots, X_t),
\end{equation}
where \( Y_t \) represents extreme weather events at time \( t \), and \( X_1, \cdots, X_t \) are input data from time \(1\) to \(t\). This sequential formulation leverages historical data and previous predictions to enhance prediction accuracy.

\subsection{Re-scaling Models}
Re-scaling techniques, especially super-resolution (SR), enhance the resolution and quality of meteorological data by generating high-resolution outputs from lower-resolution inputs, crucial for accurate climate modeling~\cite{sluiter2009interpolation, vandal2017deepsd, cheng2020reslap, watson2020investigating}. Deep learning models, particularly CNNs, effectively learn mappings between low and high-resolution data, capturing spatial and temporal patterns to improve dataset granularity. A primary application is downscaling precipitation data from coarse (e.g., 100 km) to fine resolutions (e.g., 12.5 km), better representing localized phenomena like extreme events and supporting more accurate weather predictions and climate assessments.

The objective function for re-scaling tasks typically uses MSE:
\begin{equation}
	\mathcal{L} = \text{MSE}(X_{\text{HR}}, f(X_{\text{LR},\theta})),
\end{equation}
where \(X_{\text{HR}}\) is the high-resolution groundtruth and \(X_{\text{LR}}\) the low-resolution input. For temporal data:
\begin{equation}
	\mathcal{L} = \text{MSE}(X_{\text{HR}_t}, f(X_{\text{LR}_1}, \cdots, X_{\text{LR}_t})),
\end{equation}
where \(X_{\text{HR}_t}\) is the high-resolution data at time \(t\).

More generally, re-scaling can be expressed as:
\begin{equation}\label{eq:rescale}
	Y_t \sim P(Y_t \mid X_1, \cdots, X_t),
\end{equation}
where \(Y_t\) is the re-scaled output at time \(t\), leveraging historical inputs to generate detailed future states.

\subsection{Universal Embedding}
Universal embedding techniques represent diverse meteorological data modalities within a common space, enabling integration across various meteorological-related tasks. This unified approach enhances model flexibility and adaptability~\cite{kratzert2019towards, man2023w}. Deep learning architectures, such as LSTMs and Masked AutoEncoders, have proven effective in creating universal embeddings. For example, Kratzert et al.~\cite{kratzert2019towards} used EA-LSTM to model rainfall–runoff processes, while W-MAE~\cite{man2023w} employed self-supervised pre-training on ERA5 datasets, improving meteorological variable predictions and enhancing models like FourCastNet.

The objective for universal embedding is:
\begin{equation}
	\mathcal{L} = \text{MSE}\left( [X, Y, Z, \cdots], f([X, Y, Z, \cdots],\theta) \right),
\end{equation}
where \( f([X, Y, Z, \cdots],\theta) \) reconstructs inputs from various modalities, ensuring coherent representation over time.

Extending this, we aim to learn a universal embedding for meteorological input sequences:
\begin{equation}
	E_t \sim P(E_t \mid [X, Y, Z, \cdots]_1, \cdots, [X, Y, Z, \cdots]_t),
\end{equation}
where \(E_t\) is the embedding at time \(t\), and \([X, Y, Z, \cdots]_1, \cdots, [X, Y, Z, \cdots]_t\) are inputs from various modalities. This framework integrates multiple data sources into a unified space, supporting diverse tasks like forecasting, classification, and anomaly detection.

\subsection{Relationship Mining}
Relationship mining in climate analysis explores connections between climatic factors, essential for understanding climate dynamics. Early methods used graph-based techniques to identify teleconnections~\cite{kawale2013graph}, while recent studies employ GNNs for more effective modeling~\cite{romanova2023gnn, chishtie2024advancing, bhandari2024recent}.

GNNs represent climate variables as graph nodes, capturing relationships through edges. Techniques like edge prediction and graph completion reveal hidden dependencies and predict missing links, improving climate impact assessments. The process can be formalized as:
\begin{equation}
	G'_t \sim P(G'_t \mid G_1, G_2, \cdots, G_t)
\end{equation}
where \(G'_t\) is the predicted climate graph at time \(t\), refining relationships between variables over time. This approach deepens the understanding of interactions among climate factors, such as temperature, precipitation, and wind patterns, enhancing climate modeling, prediction, and mitigation strategies through dynamic relationship mining using GNNs.

\subsection{Others}
While we cover key AI applications in weather and climate sciences, the field encompasses many additional methodologies and nuances. The diverse weather and climate phenomena and their complex spatiotemporal dynamics suggest that other approaches may further advance the field. As weather and climate sciences evolve, new techniques and models will emerge to tackle ongoing and future challenges.




\end{document}